\documentclass[10pt,twocolumn,letterpaper]{article}

\usepackage[pagenumbers]{iccv} 

\usepackage{colortbl}
\usepackage{tabulary}
\usepackage{etoolbox}
\usepackage{graphicx}
\usepackage{amsmath}
\usepackage{amssymb}
\usepackage{booktabs}
\usepackage{float}
\usepackage{xcolor}
\usepackage{mathtools, nccmath}

\usepackage{soul}
\usepackage{xspace}\xspace
\usepackage{array}
\usepackage{dblfloatfix}
\usepackage{transparent}
\usepackage{adjustbox}
\usepackage{algorithm}
\usepackage{listings}
\usepackage{arydshln}
\usepackage{bbding}
\usepackage{twemojis}
\usepackage{multirow}
\usepackage{graphicx}
\usepackage{caption}
\usepackage{subcaption}
\usepackage{algorithmic}
\usepackage{wrapfig}
\usepackage{caption}
\usepackage{CJKutf8}

\definecolor{iccvblue}{rgb}{0.21,0.49,0.74}
\usepackage[pagebackref,breaklinks,colorlinks,allcolors=iccvblue]{hyperref}

\def\paperID{7151} 
\def\confName{ICCV}
\def\confYear{2025}

\title{Leveraging the Power of MLLMs for Gloss-Free Sign Language Translation}

\author{Jungeun Kim\textsuperscript{\rm 1*}, Hyeongwoo Jeon\textsuperscript{\rm 2*}, Jongseong Bae\textsuperscript{\rm 1}, Ha Young Kim\textsuperscript{\rm 2\textdagger}\\
\textsuperscript{\rm 1}Department of Artificial Intelligence, Yonsei University\\
\textsuperscript{\rm 2}Graduate School of Information, Yonsei University\\
{\tt\small \{jekim5418, hyeong1204, js.bae, hayoung.kim\}@yonsei.ac.kr}
}

\begin{document}
\maketitle
\let\thefootnote\relax\footnotetext{\textsuperscript{\rm *}These authors contributed equally.}
\let\thefootnote\relax\footnotetext{\textsuperscript{\textsuperscript{\rm \textdagger}}Corresponding author.}
\begin{abstract}
Sign language translation (SLT) is a challenging task that involves translating sign language images into spoken language.
For SLT models to perform this task successfully, they must bridge the modality gap and identify subtle variations in sign language components to understand their meanings accurately. 
To address these challenges, we propose a novel gloss-free SLT framework called \textbf{M}ulti\textbf{m}odal \textbf{S}ign \textbf{L}anguage \textbf{T}ranslation (\textbf{MMSLT}), which leverages the representational capabilities of off-the-shelf multimodal large language models (MLLMs). 
Specifically, we use MLLMs to generate detailed textual descriptions of sign language components.
Then, through our proposed multimodal-language pre-training module, we integrate these description features with sign video features to align them within the spoken sentence space.
Our approach achieves state-of-the-art performance on benchmark datasets PHOENIX14T and CSL-Daily, highlighting the potential of MLLMs to be utilized effectively in SLT.
Code is available at \href{https://github.com/hwjeon98/MMSLT}
{{https://github.com/hwjeon98/MMSLT}}.
\end{abstract}    
\section{Introduction}
\label{sec:intro}

Sign language (SL) is the primary mode of communication for deaf individuals. It relies on visual elements such as hand gestures, body movements, and facial expressions~\cite{stokoe2005sign,sharma2020sign,rastgoo2022need}, collectively called \textit{SL components}.
The sign language translation (SLT) task aims to bridge communication gaps by converting SL videos into spoken sentences.
However, this task poses significant challenges, as the model needs to identify SL components and convert them into spoken sentences. 
Specifically, it requires overcoming the modality gap—transitioning from visual cues to text—while also understanding the intricate cross-modal relationships involved~\cite{veale1998challenges, duarte2019cross, zhao2021conditional}.

\begin{figure}[t!]
\centering
    \centerline{\includegraphics[width=\linewidth]{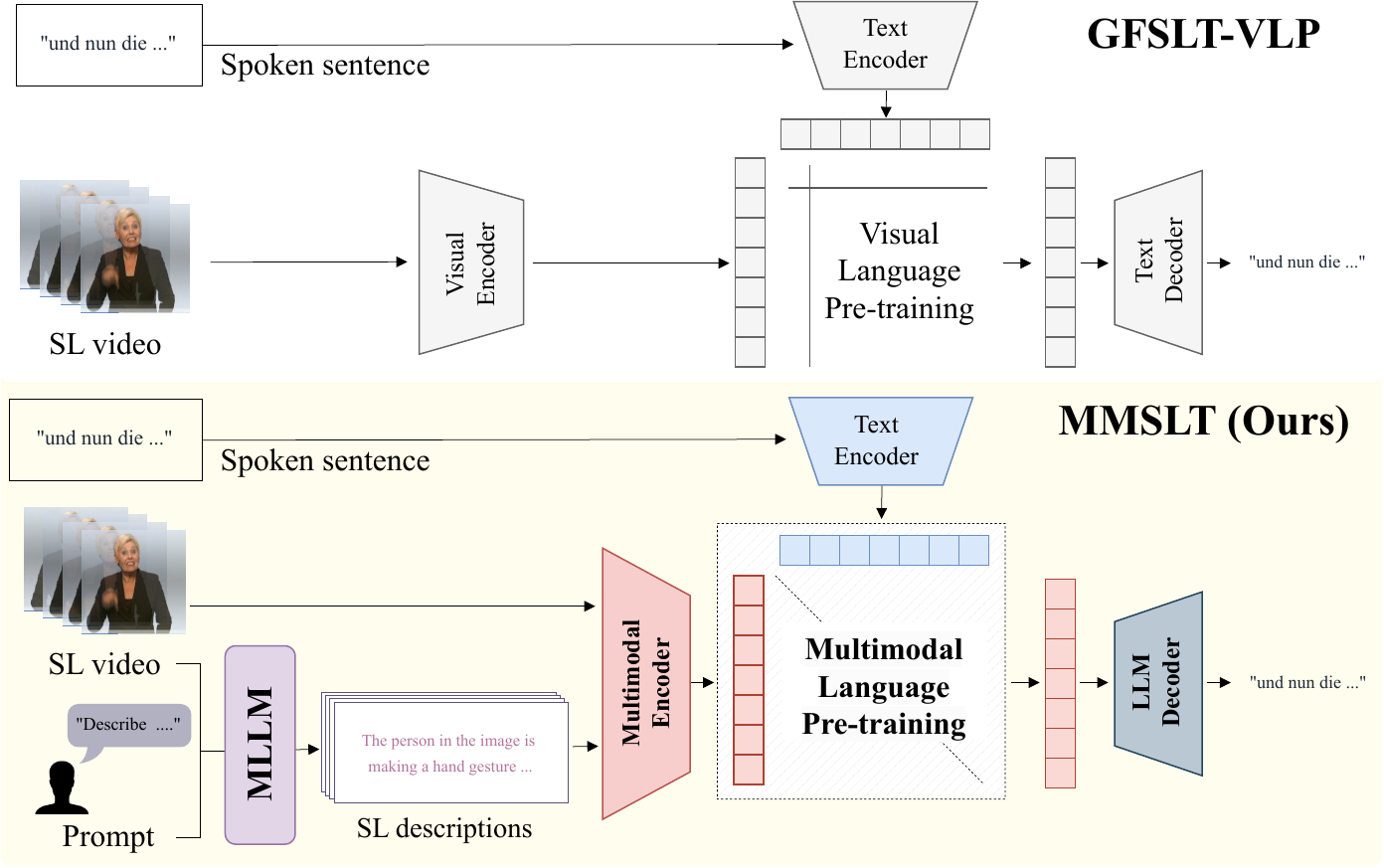}}
    \caption{Comparison of the proposed MMSLT with GFSLT-VLP~\cite{zhou2023gloss}, a representative gloss-free SLT approach. 
    Previous gloss-free SLT methods extract visual features from SL images and align them with spoken sentences.
    In contrast, our MMSLT leverages an MLLM to represent SL images as text, integrates the resulting representations with the images, and aligns the fused modalities with spoken sentences.
    }
\label{fig:comparison}
\end{figure}

To enhance SLT, gloss-based models~\cite{camgoz2018neural, voskou2021stochastic, zhang2023sltunet} that use mid-level supervision with glosses have been developed, as well as weakly supervised gloss-free models~\cite{yin2023gloss, zhao2021conditional, jiao2024visual} that indirectly leverage glosses to extract visual features through an encoder trained on SL recognition datasets.
However, these approaches still rely on labor-intensive gloss annotations, which pose scalability challenges and create information bottlenecks~\cite{hwang2024universal, hwang2024efficient}.
Therefore, recent studies have shifted towards gloss-free SLT models that directly translate SL videos into spoken language~\cite{zhou2023gloss, wong2024sign2gpt, gong2024llms}.
Most gloss-free models utilize modules such as visual encoders and text decoders to extract visual features from SL images, transforming these features into text-like representations that large language models (LLMs) can process for translation~\cite{zhou2023gloss, achiam2023gpt, gong2024llms}, as illustrated in Fig.~\ref{fig:comparison}. 
However, since these visual features are obtained from SL images, they can be affected by visual information unrelated to SLT, such as background elements or clothing colors, limiting their ability to accurately represent SL components.

To address these challenges, we pose the question, ``How can we extract the information that explains detailed SL components and is easy for LLMs to understand?" In response to this question, we propose that deriving ``SL descriptions," which are texts that describe SL components, could provide a solution.
Additionally, SL descriptions are in text form, so they help alleviate the modality gap. To this end, as shown in Fig.~\ref{fig:comparison}, we consider using a MLLM (Multimodal Large Language Model), which processes multiple modalities simultaneously to integrate information and generate text-based answers~\cite{wu2023multimodal}.

\begin{figure*} 
\centering
    \begin{subfigure}[t]{.33\textwidth}
        \centering
        {\includegraphics[width=\linewidth]{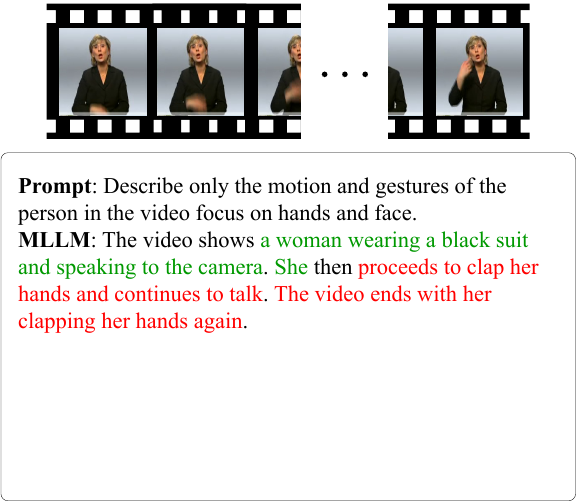}}
        \caption{Video-based MLLM with prompt type (3).}
       \label{fig:video} 
    \end{subfigure} %
    \hfill
    \begin{subfigure}[t]{.33\textwidth} 
        \centering
        \includegraphics[width=\linewidth]{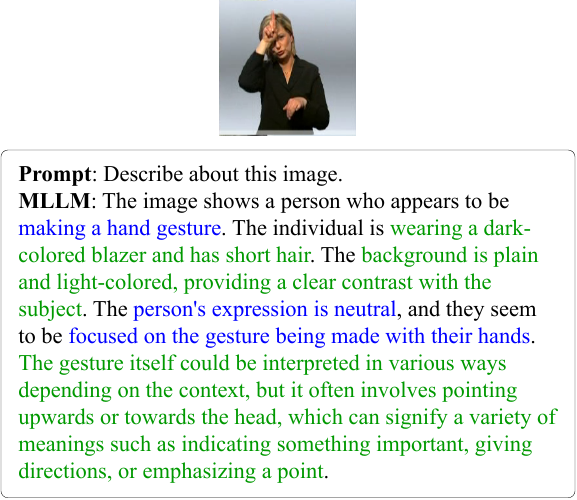}
        \caption{Image-based MLLM with prompt type (1).}
       \label{fig:simple} 
    \end{subfigure} 
    \begin{subfigure}[t]{.33\textwidth} 
        \centering
        \includegraphics[width=\linewidth]{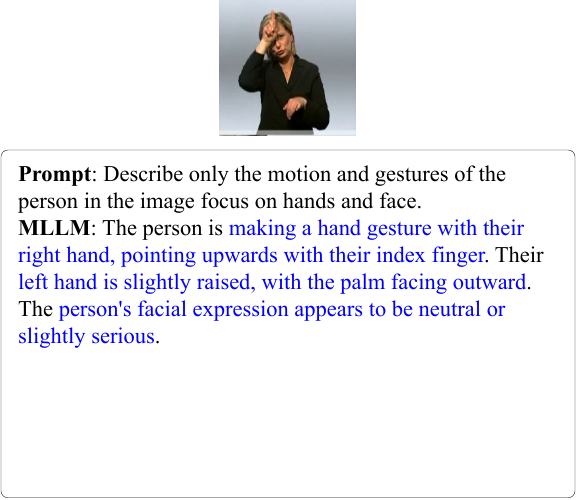}
        \caption{Image-based MLLM with prompt type (3).}
       \label{fig:main} 
    \end{subfigure} 
   \caption{Comparison of SL descriptions based on the type of MLLM and prompt. (a) shows responses from Video-LLaVA 7B~\cite{lin2023video}, while (b) and (c) display inference results for LLaVA-OneVision 7B~\cite{li2024llavaov}. 
   Incorrect parts are highlighted in red, irrelevant information is in green, and accurate SL component descriptions are in blue.
    }
    \vspace{-0.2cm}
    \label{fig:comparison_int}
\end{figure*}

MLLMs can produce appropriate responses that conform to the context and purpose of a task based on a provided image and prompt. To investigate the potential of MLLMs for SLT, we explore their capabilities by employing various types of MLLMs and applying prompt engineering to extract descriptions of SL components from images.
Our findings indicate that MLLMs can accurately describe fine SL components, such as the position of hands and fingers, while effectively minimizing extraneous details, including the signer's gender or background, as illustrated in Fig.~\ref{fig:main}. However, despite the representation power of MLLMs, there has been no prior research on their application to SLT. 
Furthermore, while MLLMs effectively describe numerous SL components, they struggle to capture certain visual information, particularly detailed facial features, such as the shapes of eyebrows and lips.

Inspired by these observations, we propose a novel gloss-free approach called \textbf{M}ulti\textbf{m}odal \textbf{S}ign \textbf{L}anguage \textbf{T}ranslation framework (\textbf{MMSLT}).
Our MMSLT consists of two main modules to enable a comprehensive understanding of SL through the complementary integration of modalities, SL descriptions and SL images, as shown in Fig.~\ref{fig:overall}. 
First, the Generating Sign Language Description via MLLM (GSD-MLLM) module serves as a pre-processing step, prompting a pre-trained MLLM to generate SL descriptions. 
Then, the Multimodal-Language Pre-training (MMLP) module integrates the two modality features and aligns them with the target spoken sentence space, bridging the modality gap and improving translation accuracy. Furthermore, to enhance efficiency, we introduce a description mapper that predicts description embeddings from the embedding features of SL images, alleviating the computational burden of using MLLM during each inference. 

To evaluate MMSLT, we perform extensive experiments on two benchmark datasets: PHOENIX14T~\cite{camgoz2018neural} and CSL-Daily~\cite{zhou2021improving}.
Results show that MMSLT significantly outperforms previous state-of-the-art (SOTA) gloss-free SLT methods, particularly on the large-scale CSL-Daily dataset.
The main contributions of MMSLT are as follows:
\begin{itemize}
    \item 
    We propose MMSLT, a gloss-free SLT framework, the first to leverage an off-the-shelf MLLM. To facilitate efficient inference without requiring the use of MLLMs, we design a description mapper module.
    
    \item 
    By analyzing various MLLMs and prompts, we propose the GSD-MLLM module as a pre-processing step, which generates detailed SL descriptions from SL images.
    \item We introduce the MMLP module, which effectively integrates two modalities, SL descriptions and SL images, and aligns them with the target sentence space to reduce the modality gap.
    
    \item We demonstrate the effectiveness of MMSLT through extensive experiments, achieving SOTA gloss-free SLT performance on two benchmark datasets. MMSLT markedly improves both BLEU-4 and ROUGE scores, indicating effective translation in complex syntax and long contexts.
\end{itemize}

\section{Related Works}

\subsection{Gloss-free Sign Language Translation}
Glosses provide intermediate supervision by expressing SL components in textual form, which is crucial for understanding SL and translating it into spoken language~\cite{camgoz2020sign, voskou2021stochastic}.
In contrast, gloss-free SLT bypasses glosses and directly translates SL videos into spoken sentences using only paired video-sentence data.
Consequently, this approach tends to have a greater modality gap than gloss-based~\cite{camgoz2020sign, voskou2021stochastic, zhou2021spatial, zhou2021improving,yin2021simulslt, jin2022prior,chen2022simple, chen2022two, zhang2023sltunet} or weakly supervised gloss-free SLT models~\cite{li2020tspnet, yin2023gloss, fu2023token, jiao2024visual}, which generally results in lower performance.
To address this gap, CSGCR~\cite{zhao2021conditional} predicts possible language words, generates sentences based on these predictions, and selects the most appropriate sentence using cosine similarity with SL images.
GFSLT-VLP~\cite{zhou2023gloss} introduces a pre-training method that aligns SL images with spoken sentences, converting them into LLM-friendly features.

Leveraging this advancement, subsequent gloss-free models focus on generating more comprehensible features for LLMs.
FLa-LLM~\cite{chen2024factorized} introduces a two-step approach: initially training on visual information from SL images using a lightweight translation model and fine-tuning the LLM for SLT.
Sign2GPT~\cite{wong2024sign2gpt} pre-trains a sign encoder by aligning visual features with prototypes under the supervision of pseudo-glosses—words from spoken sentences via Parts-of-Speech (POS) tagging (\eg, nouns and numbers)—and then utilizes it for SLT.
Lastly, SignLLM~\cite{gong2024llms} normalizes input video to generate sign tokens with linguistic features compatible with LLMs.
However, to date, no SLT study has directly represented SL images in a textual format. 
Representing the same data across different modalities, as proposed in our approach, allows the two modalities to complement each other and create synergy.
Furthermore, this approach can also help mitigate the modality gap.

\subsection{Multimodal Large Language Models}
Early MLLMs like Flamingo~\cite{alayrac2022flamingo} and BLIP-2~\cite{li2023blip2} focus on mapping language to visual meaning. However, these models are limited to merely describing images. 
Motivated by the versatility of LLMs~\cite{touvron2023llama, achiam2023gpt, taori2023alpaca}, LLaVA~\cite{liu2024visual} introduces a method to generate multimodal instruction-following data using LLMs, and then fine-tunes a large multimodal model end-to-end by connecting the visual encoder of CLIP~\cite{radford2021learning} with the language decoder of Vicuna~\cite{chiang2023vicuna}.
This approach has led to the emergence of MLLMs that enhance vision and language comprehension by applying LLMs~\cite{zhu2023minigpt, gong2023multimodal, chen2024internvlscalingvisionfoundation, wang2024qwen2vlenhancingvisionlanguagemodels, agrawal2024pixtral12b, li2024llavaov}.
Furthermore, video-specialized MLLMs have been developed by combining video-based models with LLMs, with some models capable of generating specific modalities such as images or audio~\cite{peng2023kosmos, zhang2023speechgpt}.
In addition, there has been research on the development of MLLMs specialized for specific tasks, such as motion understanding or motion tracking for skeleton data~\cite{hong2024egolm, mo2024mochat}. 
However, to the best of our knowledge, this is the first work to both extract descriptions from images using existing MLLMs and leverage these capabilities for SLT tasks.

\subsection{Vision-Language Pre-training}
Existing visual-language pre-training (VLP) models can be categorized into single-stream and dual-stream models~\cite{chen2023vlp}. 
Single-stream models~\cite{chen2020uniter, yao2021filip, wang2021simvlm} integrate visual and text modalities as input to a transformer~\cite{vaswani2017attention} encoder. 
In this framework, the relationship between the two modalities is learned through the self-attention mechanism. In contrast, dual-stream models~\cite{luo2020univl, jia2021scaling, li2021align} use separate encoders for each modality, and the relationship between the modalities is trained through contrastive learning or cross-attention. These pre-trained models have demonstrated remarkable performance in downstream tasks.


GFSLT-VLP~\cite{zhou2023gloss} is the first study to utilize CLIP~\cite{radford2021learning}, a pioneering approach in VLP, to tackle the modality gap in gloss-free SLT.
Like CLIP, GFSLT-VLP adopts a dual-stream structure, pre-training the alignment between sign videos and spoken sentences before performing the translation. 
However, this approach has limitations in providing frame-level supervision. Therefore, VAP~\cite{jiao2024visual} proposes a method to learn the alignment between frames in SL videos and words in spoken sentences. Despite these advancements, existing SLT studies using VLP have primarily focused on elucidating the relationship between a single modality (visual) and a single modality (text). 
As far as we know, no SLT research currently utilizes multimodality (visual and text) as input to learn the alignment for a single modality (text).
\section{MLLMs and Prompts: Preliminary Analysis}
\label{sec:preliminary}
\begin{figure}[t!]
\centering
    \centerline{\includegraphics[width=\linewidth, height=2.5in]{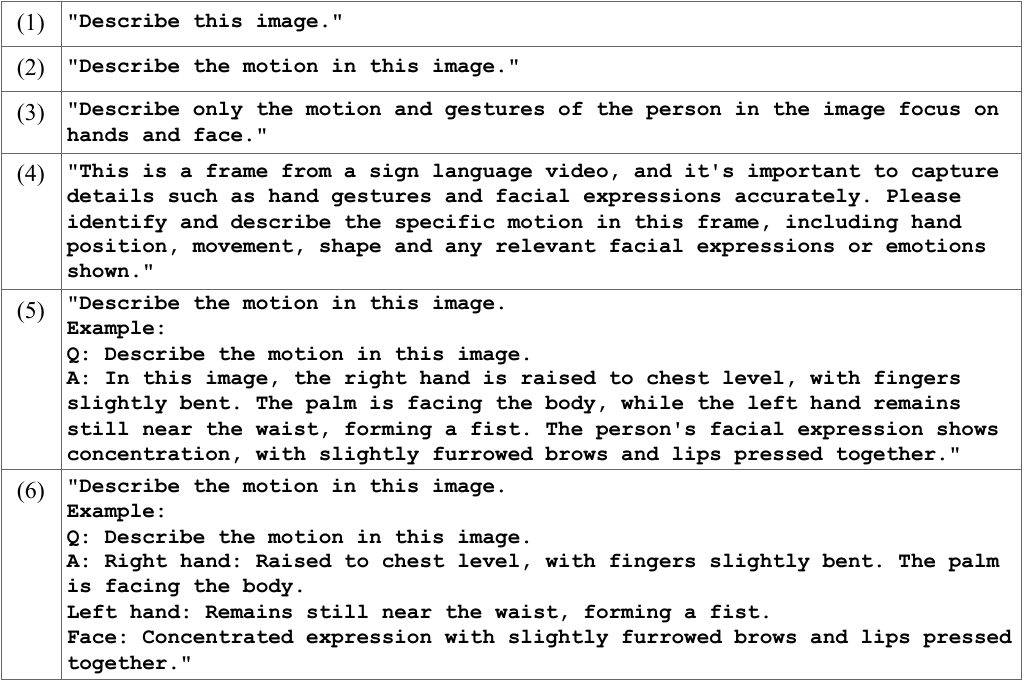}}
    \caption{List of prompts.}
\label{fig:prompt}
\vspace{-0.2cm}
\end{figure}

We explore the applicability of MLLMs to SLT using various models and prompts. The details are provided in the supplementary, with a summary as follows. \\
\\
\noindent\textbf{Video-based MLLM vs Image-based MLLM.}
We first examine the potential of video-based and image-based MLLMs for SLT by evaluating their ability to describe SL components. 
As shown in Fig.~\ref{fig:video}, video-based MLLMs, such as Video-LLaMA~\cite{damonlpsg2023videollama} and Video-LLaVA~\cite{lin2023video}, demonstrate limited capability in capturing SL components. 
Instead, they provide simplistic summaries, like `using her hands to gesture' or repeat the same content, and include general descriptions of situations unrelated to SL, such as `wearing a suit' or inaccurate information.
On the other hand, image-based MLLMs such as LLaVA-Next~\cite{li2024llavanext}, InternVL~\cite{chen2024internvlscalingvisionfoundation}, QwenVL2~\cite{wang2024qwen2vlenhancingvisionlanguagemodels}, Pixtral~\cite{agrawal2024pixtral12b}, and LLaVA-OneVision~\cite{li2024llavaov} generate detailed descriptions of SL components, such as `fingers slightly spread apart,' or `eyes are focused and directed toward.' Notably, as shown in Fig.~\ref{fig:main}, LLaVA-OneVision provides descriptions focused solely on SL components without including unrelated information. 
In this experiment, large models, such as LLaVA-OneVision 72B, are excluded from the analysis due to the processing time and resource constraints associated with generating SL descriptions.\\
\\
\noindent \textbf{Prompt Engineering.}
To effectively leverage MLLMs, it is crucial to select an appropriate model and formulate prompts that explicitly delineate the information to be extracted or provide contextual clarity for the content.
To achieve this, we perform inference with six distinct prompts, as shown in Fig.~\ref{fig:prompt}.
These can be grouped into simple (1, 2), detailed (3, 4), and in-context (5, 6) types.
Prompts (1), (2), and (4) elicit responses that primarily focus on irrelevant information, as shown in Fig.~\ref{fig:simple}, while prompts (3), (5), and (6) generate responses more concentrated on SL components. 
However, responses to prompt (6) contain inaccuracies, while the responses to prompt (5) are more accurate and provide more detailed information.
Nevertheless, they often reuse example sentences and produce repetitive answers across consecutive frames.
In contrast, as shown in Fig.~\ref{fig:main}, prompt (3) provides detailed descriptions of SL components, such as hand shape, signer’s gaze, and mouth shape, effectively conveying the meanings of the signer’s gestures and facial expressions. Furthermore, it avoids referencing external information irrelevant to understanding SL and successfully captures subtle differences between consecutive sign gestures.

\section{Method}
\label{sec:method}
\begin{figure*}[ht]
    \centerline{\includegraphics[width=1\linewidth, height=2.4in]{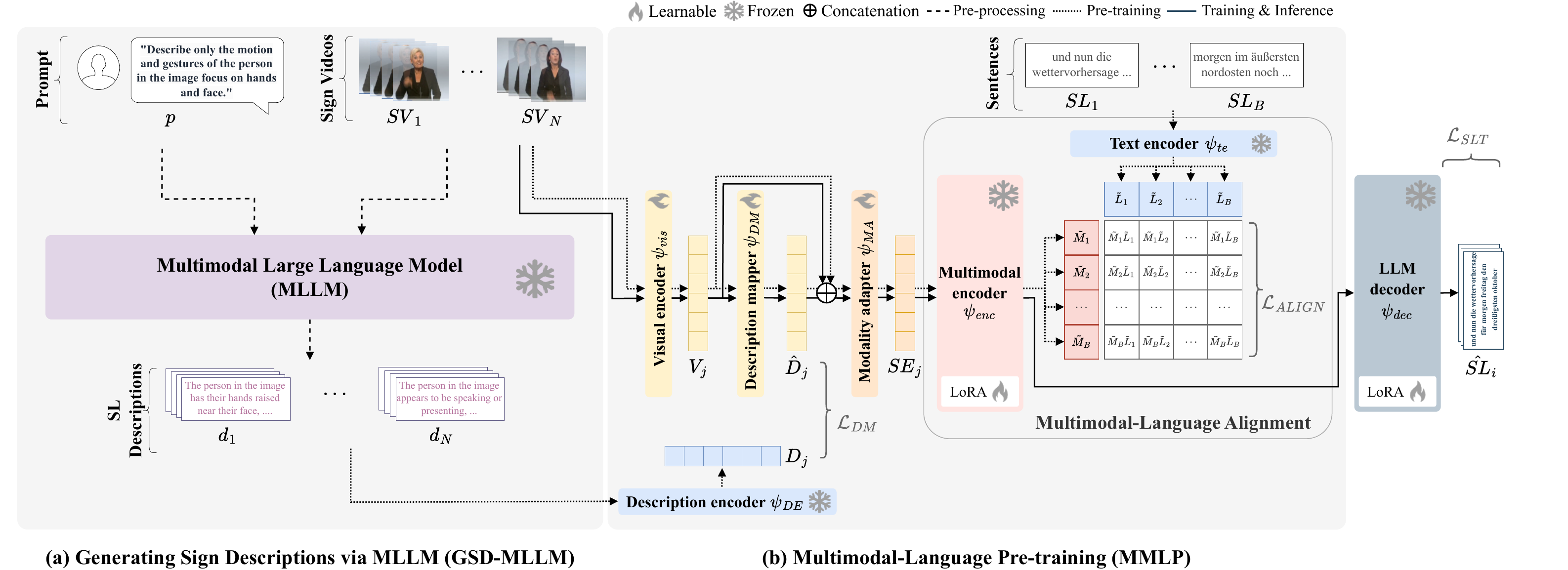}}
    \caption{MMSLT overview. MMSLT comprises two modules: \textbf{(a) Generating Sign Descriptions via MLLM (GSD-MLLM)}: This module utilizes prompts along with SL videos as inputs to the MLLM, generating SL descriptions that describe the SL components (dashed line in the figure). \textbf{(b) Multimodal Sign Language Pre-training (MMLP)}: In this module, sign videos and SL descriptions are integrated and aligned with spoken sentences. 
    We introduce a description mapper, which allows efficient inference without requiring the MLLM by approximating the embedding features of SL descriptions based on the SL video (solid line in the figure).
    Following the MMLP, pre-trained networks are fine-tuned, and the initialized LLM decoder is trained for SLT.}
\label{fig:overall}
\vspace{-0.2cm}
\end{figure*}
This section introduces our proposed MMSLT framework. 
We begin with a detailed explanation of how SL descriptions are generated using MLLM in Sec.~\ref{subsec:GSD-MLLM}. 
Next, Sec.~\ref{subsec:MMLP} describes the MMLP module, which integrates SL images and the corresponding SL descriptions while aligning them with the target spoken sentences. 
Finally, Sec.~\ref{subsec:SLT} explains the training process of SLT.

\subsection{Generating SL Descriptions via MLLM}
\label{subsec:GSD-MLLM}
Based on the analysis in Sec.~\ref{sec:preliminary}, we utilize an image-based MLLM with a prompt $p$ to generate a set of SL descriptions $d_j = \{d_{j,t}\}_{t=1}^T$ for the $j$-th SL video ${SV}_j\in\mathbb{R}^{T \times H \times W}$, as described in Fig.~\ref{fig:overall}. In this context, the total number of SL videos is $N$, the height and width of the frames are $H$ and $W$, respectively. The total number of frames is $T$.

\subsection{Multimodal-Language Pre-training}
\label{subsec:MMLP}
To facilitate synergy between two modalities, we integrate SL images into the SL descriptions.
It is evident that a modality gap exists between SL videos and spoken sentences. 
To bridge this gap, we propose MMLP, a pre-training module that effectively fuses SL videos with their corresponding SL descriptions and learns to align them with spoken sentences, as illustrated in Fig.~\ref{fig:overall}. \\
\\
\noindent \textbf{Description Mapper.} 
First, we extract the \textit{visual feature} $V_j \in\mathbb{R}^{T \times C}$ from the $j$-th SL video ${SV}_j$ within a mini-batch of size $B$ using the image encoder $\psi_{vis}$, which is a ResNet18~\cite{he2015deepresiduallearningimage} pre-trained on ImageNet~\cite{5206848}, where $C$ represents the dimensionality of $V_j$.
Simultaneously, we obtain the \textit{description embedding feature} 
$D_{j,t} \in\mathbb{R}^{1 \times \Bar{C}}$ from the SL description of each frame $d_{j,t}$ using the description encoder $\psi_{de}$, a pre-trained, frozen 12-layer BERT~\cite{devlin2019bertpretrainingdeepbidirectional}. 
In this context, we define the embedding feature of the \texttt{[CLS]} token as $D_{j,t}$, encoding sentence-level information, with $\Bar{C}$ denoting the dimensionality of $D_{j,t}$.
These description embedding features collectively form the \textit{description feature} $D_j=\{D_{j,t}\}_{t=1}^T$.
However, if $D_j$ is directly utilized to integrate SL videos and SL descriptions, generating SL descriptions via MLLM becomes necessary during the inference. 
This necessity leads to increased computational costs and extended inference times.
To address this issue, we propose a description mapper $\psi_{dm}$ characterized by a simple two-layer MLP structure that predicts $D_j$ from $V_j$. 
This also helps bridge the modality gap between the input data sources and allows us to obtain the \textit{approximated description feature} $\hat{D}_j$, which can be expressed as follows:
\begin{equation}
     \hat{D}_j = \psi_{dm}(V_j), \, V_j=\psi_{vis}({SV}_j).
    \label{eq:description_mapping}
\end{equation}
To minimize $\hat{D}_j$ and $D_j$, we define the loss function as follows:
\begingroup
\small
 \begin{equation}
    \mathcal{L}_{\text{DM}} = \frac{1}{B}  \frac{1}{T} \sum_{j=1}^{B} \sum_{t=1}^{T} \left\| \hat{D}_{j,t} - D_{j,t} \right\|_2^2.
\end{equation}
\endgroup
\\
\noindent \textbf{Modality Adapter.} 
To combine the visual feature and the approximated SL description feature, we introduce a modality adapter $\psi_{ma}$ consisting of two successive 1D convolution and max-pooling layers, followed by a two-layer MLP.
Since SL typically consists of multiple frames, and $V_j$ and $\hat{D}_j$ are derived from consecutive SL frames, we employ a 1D convolution for temporal modeling consistent with previous studies~\cite{zhou2023gloss, chen2024factorized, wong2024sign2gpt}. 
Additionally, MLP layers are employed to integrate information from both modalities.
The captured $V_j$ and $\hat{D}_j$ are concatenated and then input into the modality adapter, resulting in the extraction of the \textit{sign element feature} ${SE}_j \in\mathbb{R}^{T^{\prime} \times C^{\prime}}$, where $T^{\prime}$ is the reduced sequence length, and $C^{\prime}$ is the embedding dimension.
This process can be expressed as follows:
\begin{equation}
    SE_j = \psi_{ma}(V_j \oplus \hat{D}_j)
    \label{eq:concat},
\end{equation}
where $\oplus$ denotes the concatenation operation. \\
\\
\noindent \textbf{Multimodal Encoder with LoRA.} 
To extract the long-term representation of ${SE}_i$~\cite{zhou2023gloss}, we introduce a multimodal encoder $\psi_{enc}$.
For this purpose, we utilize the mBART encoder~\cite{tang2020multilingualtranslationextensiblemultilingual}, which consists of 12 layers and is initialized with parameters pre-trained on a large corpus. To facilitate adaptation to the SL dataset while preserving the pre-trained knowledge, we apply the LoRA~\cite{hu2022lora} technique, which enables parameter-efficient fine-tuning by adding low-rank matrices into the weight matrix of the LLM. The resulting output, referred to as the \textit{multimodal visual-textual feature (multimodal feature)}, is denoted as ${M}_j \in \mathbb{R}^{T^{\prime} \times C^{\prime}}$ and can be represented as follows:
\begin{equation}
    {M}_j = \psi_{enc}({SE}_j).
    \label{eq:llm encoder}
\end{equation}
\noindent \textbf{Multimodal-Language Alignment.}
${M}_j$ encompasses visual and textual information; however, a modality gap still remains with the target spoken sentence. 
To tackle this challenge and align ${M}_j$ with the target spoken sentence, we present a multimodal-language alignment method.

Let ${SL}_j$ be the target spoken sentence corresponding to ${SV}_j$. 
We embed ${SL}_j$ into text features $L_j \in \mathbb{R}^{\Bar{T} \times C^{\prime}}$, where $\Bar{T}$ denotes the number of tokens in the spoken sentence, using a text encoder $\psi_{te}$, a frozen 12-layer mBART encoder~\cite{tang2020multilingualtranslationextensiblemultilingual} pre-trained on a large corpus, similar to the $\psi_{enc}$.
Since our objective is to align $\{M_j, L_j\}_{j=1}^{B}$, we apply contrastive learning, which is consistent with previous studies~\cite{zhou2023gloss, jiao2024visual}. 
Specifically, we perform average pooling over the frame sequence $T^{\prime}$ and the token sequence $\Bar{T}$ to obtain the \textit{global multimodal feature} $\tilde{M}_j \in \mathbb{R}^{C^{\prime}}$ and \textit{global spoken-sentence feature} $\tilde{L}_j \in \mathbb{R}^{C^{\prime}}$.
Finally, we align the pairs $\{\tilde{M}_j, \tilde{L}_j\}_{j=1}^{B}$ using the loss function as follows:
\begingroup
\small
\begin{equation}
\begin{aligned}
    \mathcal{L}_{\text{ALIGN}} &= - \frac{1}{2B}( \sum_{j=1}^{B} \log \frac{\exp(sim(\tilde{M}_j, \tilde{L}_j) / \tau)}{\sum_{k=1}^{B} \exp(sim(\tilde{M}_j, \tilde{L}_k) / \tau)} \\
    & \quad + \sum_{j=1}^{B} \log \frac{\exp(sim(\tilde{L}_j, \tilde{M}_j) / \tau)}{\sum_{k=1}^{B} \exp(sim(\tilde{L}_j, \tilde{M}_k) / \tau)}),
\end{aligned}
\end{equation}
\endgroup
where $sim(x, y)$ represents the cosine similarity between $x$ and $y$, and $\tau$ is a learnable temperature parameter.\\

Thus, we define the final loss function in the MMLP as follows:
\begin{equation}
    \mathcal{L}_{\text{MMLP}} = \mathcal{L}_{\text{ALIGN}} + \lambda\mathcal{L}_{\text{DM}},
\end{equation}
where $\lambda$ is a hyperparameter that controls the weight of $\mathcal{L}_{\text{ALIGN}}$ and $\mathcal{L}_{\text{DM}}$. 
\definecolor{lavender}{rgb}{0.9, 0.9, 0.98}
\begin{table*}[ht!]
    \centering
    \setlength{\tabcolsep}{4pt}
    \renewcommand{\arraystretch}{1}
    \scriptsize
        \begin{tabular}{l|cccccccccc}
        \hline
          & \multicolumn{5}{c|}{Dev} & \multicolumn{5}{c}{Test} \\
        Method &  BLEU-1 & BLEU-2 & BLEU-3 & BLEU-4 & \multicolumn{1}{c|}{ROUGE} & BLEU-1 & BLEU-2 & BLEU-3 & BLEU-4 & ROUGE \\ \hline
        \rowcolor{lavender}
        \multicolumn{11}{c}{Gloss-based}        \\                                                                                                          
        \multicolumn{1}{l|}{SLRT~\cite{camgoz2020sign}(CVPR20)}                  & 47.26  & 34.40  & 27.05  & 22.38  & \multicolumn{1}{c|}{-}     & 46.61  & 33.73  & 26.19  & 21.32  & -     \\
        \multicolumn{1}{l|}{STN-SLT~\cite{voskou2021stochastic}(ICCV21)}         & 49.12  & 36.29  & 28.34  & 23.23  & \multicolumn{1}{c|}{-}     & 48.61  & 35.97  & 28.37  & 23.65  & -     \\
        \multicolumn{1}{l|}{STMC-T~\cite{zhou2021spatial}(TMM21)}               & 47.60  & 36.43  & 29.18  & 24.09  & \multicolumn{1}{c|}{48.24} & 46.98  & 36.09  & 28.70  & 23.65  & 46.65 \\
        \multicolumn{1}{l|}{BN-TIN-Transf.+SignBT~\cite{zhou2021improving}(CVPR21)} & 51.11  & 37.90  & 29.80  & 24.45  & \multicolumn{1}{c|}{50.29} & 50.80  & 37.75  & 29.72  & 24.32  & 49.54 \\
        \multicolumn{1}{l|}{MMTLB~\cite{chen2022simple}(CVPR22)}                 & 53.95  & 41.12  & 33.14  & 27.61  & \multicolumn{1}{c|}{53.10} & 53.97  & 41.75  & 33.84  & 28.39  & 52.65 \\
        \multicolumn{1}{l|}{TS-SLT~\cite{chen2022two}(NeurIPS22)}                & 54.32  & 41.99  & 34.15  & 28.66  & \multicolumn{1}{c|}{54.08} & 54.90  & 42.43  & 34.46  & 28.95  & 53.48 \\
        \multicolumn{1}{l|}{SLTUNET~\cite{zhang2023sltunet}(ICLR23)}             & -      & -      & -      & 27.87  & \multicolumn{1}{c|}{52.23} & 52.92  & 41.76  & 33.99  & 28.47  & 52.11 \\ \hline
        \rowcolor{lavender}
        \multicolumn{11}{c}{Weakly supervised gloss-free} \\ 
        \multicolumn{1}{l|}{TSPNet~\cite{li2020tspnet}(NeurIPS20)}                & -      & -      & -      & -      & \multicolumn{1}{c|}{-}     & 36.10  & 23.12  & 16.88  & 13.41  & 34.96 \\
        \multicolumn{1}{l|}{GASLT~\cite{yin2023gloss}(CVPR23)}                 & -      & -      & -      & -      & \multicolumn{1}{c|}{-}     & 39.07  & 26.74  & 21.86  & 15.74  & 39.86 \\
        \multicolumn{1}{l|}{ConSLT~\cite{fu2023token}(ICASSP23)}                & -      & -      & -      & 21.11  & \multicolumn{1}{c|}{47.74} & -      & -      & -      & 21.59  & 47.69 \\
        \multicolumn{1}{l|}{VAP~\cite{jiao2024visual}(ECCV24)}                & 52.78      & -      & -      & 26.62  & \multicolumn{1}{c|}{51.47} & 53.07      & -      & -      & 26.16  & 51.28 \\\hline
        \rowcolor{lavender}
        \multicolumn{11}{c}{Gloss-free} \\ 
        \multicolumn{1}{l|}{NSLT~\cite{camgoz2018neural}(CVPR18)}             & 28.10  & 16.81  & 11.82  & 9.12  & \multicolumn{1}{c|}{31.00} & 27.10  & 15.61  & 10.82  & 8.35  & 29.70 \\
        \multicolumn{1}{l|}{NSLT+Bahdanau~\cite{camgoz2018neural, bahdanau2016neuralmachinetranslationjointly}(CVPR18)}                      & 31.87  & 19.11  & 13.16  & 9.94  & \multicolumn{1}{c|}{31.80} & 32.24  & 19.03  & 12.83  & 9.58  & 31.80 \\
        \multicolumn{1}{l|}{NSLT+Luong~\cite{camgoz2018neural,luong-etal-2015-effective}(CVPR18)}             & 31.58  & 18.98  & 13.22  & 10.00  & \multicolumn{1}{c|}{32.60} & 29.86  & 17.52  & 11.96  & 9.00  & 30.70 \\
        \multicolumn{1}{l|}{CSGCR~\cite{zhao2021conditional}(TMM21)}             & 35.85  & 24.77  & 18.65  & 15.08  & \multicolumn{1}{c|}{38.96} & 36.71  & 25.40  & 18.86  & 15.18  & 38.85 \\
        \multicolumn{1}{l|}{GFSLT-VLP~\cite{zhou2023gloss}(ICCV23)}             & 44.08  & 33.56  & 26.74  & 22.12  & \multicolumn{1}{c|}{43.72} & 43.71  & 33.18  & 26.11  & 21.44  & 42.49 \\
        
        \multicolumn{1}{l|}{Sign2GPT~\cite{wong2024sign2gpt}(ICLR24)}              & -      & -      & -      & -      & \multicolumn{1}{c|}{-}     & \textbf{49.54}  & \underline{35.96}  & \underline{28.83}  & 22.52  & \textbf{48.90} \\
        \multicolumn{1}{l|}{FLa-LLM~\cite{chen2024factorized}(LREC-COLING24)}             & -      & -      & -      & -      & \multicolumn{1}{c|}{-}     & 46.29  & 35.33  & 28.03  & 23.09  & 45.27 \\
        \multicolumn{1}{l|}{SignLLM~\cite{gong2024llms}(CVPR24)}               & \underline{46.88}  & \underline{36.59}  & \underline{29.91}  & \underline{25.25}  & \multicolumn{1}{c|}{\underline{47.23}} & 45.21  & 34.78  & 28.05  & \underline{23.40}  & 44.49 \\
        \hline
        \multicolumn{1}{l|}{\textbf{MMSLT (Ours)}}          &  \textbf{48.73}  &  \textbf{37.78} &   \textbf{30.51}   &    \textbf{25.47}    & \multicolumn{1}{c|}{\textbf{48.58}}    &    \underline{48.92}     &    \textbf{38.12}    &    \textbf{30.79}    &    \textbf{25.73}    &   \underline{47.97}    \\ \hline
        \end{tabular}
    \caption{Experimental results on PHOENIX14T dataset. The best results for gloss-free models are highlighted in bold, while the second-best results are underlined.}
    \label{tbl:phoenix}
    \vspace{-0.2cm}
\end{table*}

\subsection{Sign Language Translation}
\label{subsec:SLT}
To perform end-to-end gloss-free SLT, we first inherit the pre-trained networks from the MMLP.
In this context, the description encoder $\psi_{de}$ is not utilized, and the description mapper $\psi_{dm}$ remains frozen, as it is solely employed for predicting SL descriptions. 
Hence, we fine-tune the visual encoder $\psi_{vis}$, the modality adapter $\psi_{ma}$, and the multimodal encoder $\psi_{enc}$.
Given a SL video ${SV}_i$, we extract the multimodal feature $\tilde{M}_i$. 
Subsequently, the initialized LLM decoder $\psi_{dec}$, which is structured as a 12-layer mBART decoder~\cite{tang2020multilingualtranslationextensiblemultilingual}, takes the multimodal feature $\tilde{M}_i$ as input to produce the predicted spoken sentence $\hat{SL}_i=(\hat{SL}_{i,1}, \dots, \hat{SL}_{i,\Bar{T}})$.
In this process, $\psi_{dec}$ uses an autoregressive approach, initiating the translation with the special start token \texttt{<BOS>} and generating words sequentially, until the end-of-sequence token \texttt{<EOS>} marks the end of the sentence generation.
We train the model to minimize the cross-entropy loss between the predicted tokens, denoted as $\hat{SL}_{i,j}$, and the ground truth tokens ${SL}_{i,j}$, defined as follows:
\begingroup
\small
\begin{equation}
    \mathcal{L}_{SLT} = - \sum_{j=1}^{\Bar{T}} \text{log}\, p(\hat{SL}_{i,j}|{SL}_{i,1:j-1}, {SV}_i).
\end{equation}
\endgroup
\section{Experiments}
\label{sec:experiment}
\begin{table*}[ht!]
    \centering
    \setlength{\tabcolsep}{4pt}
    \renewcommand{\arraystretch}{1}
    \scriptsize
        \begin{tabular}{l|ccccc|ccccc}
        \hline
          & \multicolumn{5}{c|}{Dev} & \multicolumn{5}{c}{Test} \\
        Method &  BLEU-1 & BLEU-2 & BLEU-3 & BLEU-4 & ROUGE & BLEU-1 & BLEU-2 & BLEU-3 & BLEU-4 & ROUGE \\ \hline
        \rowcolor{lavender}
        \multicolumn{11}{c}{Gloss-based}        \\    
        \multicolumn{1}{l|}{SLRT~\cite{camgoz2020sign}(CVPR20)}                  & 37.47  & 24.67  & 16.86  & 11.88  & \multicolumn{1}{c|}{37.96} & 37.38  & 24.36  & 16.55  & 11.79  & 36.74 \\
        \multicolumn{1}{l|}{BN-TIN-Transf.+SignBT~\cite{zhou2021improving}(CVPR21)} & 51.46  & 37.23  & 27.51  & 20.80  & \multicolumn{1}{c|}{49.49} & 51.42  & 37.26  & 27.76  & 21.34  & 49.31 \\
        \multicolumn{1}{l|}{MMTLB~\cite{chen2022simple}(CVPR22)}                 & 53.81  & 40.84  & 31.29  & 24.42  & \multicolumn{1}{c|}{53.38} & 53.31  & 40.41  & 30.87  & 23.92  & 53.25 \\
        \multicolumn{1}{l|}{TS-SLT~\cite{chen2022two}(NeurIPS22)}                & 55.21  & 42.31  & 32.71  & 25.76  & \multicolumn{1}{c|}{55.10} & 55.44  & 42.59  & 32.87  & 25.79  & 55.72 \\
        \multicolumn{1}{l|}{SLTUNET~\cite{zhang2023sltunet}(ICLR23)}               & -      & -      & -      & 23.99  & \multicolumn{1}{c|}{53.58} & 54.98  & 41.44  & 31.84  & 25.01  & 54.08 \\ \hline
        \rowcolor{lavender}
        \multicolumn{11}{c}{Weakly supervised gloss-free} \\ 
        \multicolumn{1}{l|}{TSPNet$^{\ast}$~\cite{li2020tspnet}(NeurIPS20)}                & -      & -      & -      & -      & \multicolumn{1}{c|}{-}     & 17.09  & 8.98   & 5.07   & 2.97   & 18.38 \\
        \multicolumn{1}{l|}{GASLT~\cite{yin2023gloss}(CVPR23)}                 & -      & -      & -      & -      & \multicolumn{1}{c|}{-}     & 19.00  & 9.94   & 5.98   & 4.07   & 20.35 \\
        \multicolumn{1}{l|}{ConSLT~\cite{fu2023token}(ICASSP23)}                & -      & -      & -      & 14.8   & \multicolumn{1}{c|}{41.46} & -      & -      & -      & 14.53  & 40.98 \\
        \multicolumn{1}{l|}{VAP$^{\ddagger}$~\cite{jiao2024visual}(ECCV24)}                & 50.41      & -      & -      & 21.16  & \multicolumn{1}{c|}{48.72} & 49.99      & -      & -      & 20.85  & 48.56 \\\hline
        \rowcolor{lavender}
        \multicolumn{11}{c}{Gloss-free} \\ 
        \multicolumn{1}{l|}{SLRT$^{\star}$~\cite{camgoz2020sign}(CVPR20)}            & 21.03  & 9.97  & 5.96  & 4.04  & \multicolumn{1}{c|}{20.51} & 20.00  & 9.11  & 4.93  & 3.03  & 19.67 \\
        \multicolumn{1}{l|}{NSLT+Luong~\cite{camgoz2018neural}(CVPR18)}            & 34.22  & 19.72  & 12.24  & 7.96  & \multicolumn{1}{c|}{34.28} & 34.16  & 19.57  & 11.84  & 7.56  & 34.54 \\
        \multicolumn{1}{l|}{GFSLT-VLP~\cite{zhou2023gloss}(ICCV23)}            & 39.20  & 25.02  & 16.35  & 11.07  & \multicolumn{1}{c|}{36.70} & 39.37  & 24.93  & 16.26  & 11.00  & 36.44 \\
        \multicolumn{1}{l|}{FLa-LLM~\cite{chen2024factorized}(LREC-COLING24)}         & -      & -      & -      & -      & \multicolumn{1}{c|}{-}     & 37.13  & 25.12  & 18.38  & 14.20  & 37.25 \\
        \multicolumn{1}{l|}{Sign2GPT~\cite{wong2024sign2gpt}(ICLR24)}           & -      & -      & -      & -      & \multicolumn{1}{c|}{-}     & \underline{41.75}  & \underline{28.73}  & \underline{20.60}  & 15.40  & \underline{42.36} \\
        \multicolumn{1}{l|}{SignLLM~\cite{gong2024llms}(CVPR24)}               & \underline{42.45}  & \underline{26.88}  & \underline{17.90}  & \underline{12.23}  & \multicolumn{1}{c|}{\underline{39.18}} & 39.55  & 28.13  & 20.07  & \underline{15.75}  & 39.91 \\ 
        \hline
        \multicolumn{1}{l|}{\textbf{MMSLT (Ours)}}         &   \textbf{50.05}   &   \textbf{36.39}    &   \textbf{26.91}     &    \textbf{20.51} & \multicolumn{1}{c|}{\textbf{48.53}} &  \textbf{49.87}     &   \textbf{36.37}     &   \textbf{27.29}     &    \textbf{21.11}    &  \textbf{48.92}    \\ \hline
        \end{tabular}
    \caption{Experimental results on CSL-Daily dataset. 
    In VAP~\cite{jiao2024visual}(marked with $\ddagger$), for a fair comparison, we present the performance without post-processing to convert punctuation to the Chinese format ~\cite{min2023faithful} \begin{CJK*}{UTF8}{gbsn}(\textit{e}.\textit{g}.,。,？)\end{CJK*} using Python's replace function.
    ${\ast}$, ${\star}$ represent the results reproduced by~\cite{yin2023gloss} and~\cite{zhou2023gloss}, respectively.} 
    \label{tbl:csl-daily}
    \vspace{-0.2cm}
\end{table*}

\subsection{Datasets, Metrics, and Implementation Details}
\textbf{Datasets.}
Following prior studies~\cite{zhou2023gloss, chen2024factorized, wong2024sign2gpt, gong2024llms}, we conduct experiments on two SLT benchmark datasets, PHOENIX14T~\cite{camgoz2018neural} and CSL-Daily~\cite{zhou2021improving} evaluating our method on their development (dev) and test sets.
\textbf{\textit{PHOENIX14T}} is a German SL dataset comprising three years of daily news and weather forecasts, divided into train, dev, and test sets containing 7,096, 519, and 642 videos, respectively.
The German spoken sentences have a vocabulary size of 2,887. 
\textbf{\textit{CSL-Daily}} is a large-scale continuous SLT dataset for Chinese SL, covering various daily scenarios.
The train, dev, and test sets contain 18,401, 1,077, and 1,176 videos, respectively, with a vocabulary size of 2,343 for Chinese spoken sentences. \\
\noindent \textbf{Evaluation Metrics.} Following previous studies~\cite{zhou2023gloss, gong2024llms}, we use BLEU~\cite{papineni2002bleu} and ROUGE-L (ROUGE)~\cite{lin2004rouge} scores to evaluate SLT. BLEU-n measures the precision of n-grams by computing the geometric mean of their precision scores. 
ROUGE calculates the F1 score based on the longest common subsequence between the translated output and reference translations, considering both word order and positional relationships.\\
\noindent \textbf{Implementation Details.} Please refer to supplementary.

\subsection{Comparison with State-of-the-Art}
\textbf{Results on PHOENIX14T.}
Tab.~\ref{tbl:phoenix} presents a comparison of gloss-based, weakly supervised gloss-free, and gloss-free methods on the PHOENIX14T dataset. Our MMSLT outperforms existing gloss-free SLT models, achieving the best or second-best performance across all metrics.
Notably, the BLEU-4 score on the test set, a main metric for translation quality, establishes a new SOTA for gloss-free SLT, with a substantial increase of +2.33 over the previous best, SignLLM~\cite{gong2024llms}.
Furthermore, unlike recent SOTA models, our model significantly improves both BLEU-4 and ROUGE scores, indicating its ability to understand and capture long phrases and context effectively.
Additionally, it achieves competitive performance compared to both gloss-based and weakly supervised gloss-free models. \\
\noindent \textbf{Results on CSL-Daily.}
Tab.~\ref{tbl:csl-daily} presents the results of the large-scale SLT benchmark dataset, CSL-Daily. Our model achieves SOTA performance across all metrics, demonstrating significant improvements of +5.36 in BLEU-4 and +6.56 in ROUGE scores over SignLLM~\cite{gong2024llms} and Sign2GPT~\cite{wong2024sign2gpt} on the test set, respectively. Surprisingly, it outperforms VAP~\cite{jiao2024visual}, the SOTA weakly supervised gloss-free model, across all metrics. 
This improvement may be attributed to MLLM-generated SL descriptions, which capture SL component details more accurately in the high-resolution CSL-Daily dataset.
A comparison of SL descriptions between PHOENIX14T and CSL-Daily is provided in the supplementary.
These findings highlight the effectiveness of the synergy between MLLM-generated SL descriptions and SL images in improving translation performance.
\subsection{Ablation studies}
In this section, we conduct all ablation experiments using the test set of PHOENIX14T. For further ablation studies, please refer to the supplementary materials. \\
\\
\noindent\textbf{Impact of Key Components in MMSLT.} We conduct experiments to assess the influence of the key components constituting MMSLT, including the GSD-MLLM module, the multimodal-language alignment, and the description mapper. 
Tab.~\ref{tbl:add} presents the performance achieved by incrementally adding each component.
A comparison between the first row, which translates using only SL images, and the second row, which uses both SL images and SL descriptions, indicates that the second row exhibits lower performance. 
This can be attributed to the modality gap between the SL images and the corresponding SL descriptions.
However, performance is substantially improved in the fourth row, where this gap is addressed through the multimodal-language alignment.
In addition, a comparison of the third and fourth rows highlights the effectiveness of GSD-MLLM when paired with M-L Align, as it surpasses M-L Align alone.
Moreover, comparing the second and fifth rows demonstrates a significant performance gain when employing a description mapper that mitigates the input modality gap, even without alignment. 
This highlights the importance of effectively utilizing modules capable of addressing the modality gap when incorporating SL descriptions.
Finally, a comparison between the fourth and final rows shows improved performance with the addition of the description mapper. 
This suggests that when combined with other modules, the description mapper creates a synergistic effect.
These findings underscore that each component of MMSLT plays a significant role in contributing to performance improvement.
\begin{table}[h!]
    \centering
    \scriptsize
    \setlength{\tabcolsep}{3pt}
    \renewcommand{\arraystretch}{1.1}
    \begin{tabular}{l|ccc|ccccc}
    \hline
    &GSD-MLLM & M-L Align & DM & B-1 & B-2 & B-3 & B-4 & R\\
     \hline
    (1) & -   & - & -  & 31.47  & 21.70  & 16.39  & 13.19 & 30.72 \\
    (2) & \scriptsize{\CheckmarkBold} & - & - &   30.05   &    20.89    &   15.77     &   12.78  & 29.56  \\  
    (3) & - & \scriptsize{\CheckmarkBold} & - &   47.29   &    36.43    &   28.96     &   23.77  & 46.53  \\  
    (4) & \scriptsize{\CheckmarkBold}   & \scriptsize{\CheckmarkBold} & - &    48.56    &  37.70  &    30.23    &   25.04  &  47.76  \\
    (5) & \scriptsize{\CheckmarkBold} & - & \scriptsize{\CheckmarkBold}  &    47.48    &    36.58    &   29.28     &    24.21 &  47.03    \\ 
    (6) & \scriptsize{\CheckmarkBold} & \scriptsize{\CheckmarkBold} & \scriptsize{\CheckmarkBold} &    \textbf{48.92}   &  \textbf{38.12}      &   \textbf{30.79}     &     \textbf{25.73} & \textbf{47.97}
    \\ \hline
    \end{tabular}
    \caption{Ablation study on key elements in MMSLT. B, R, M-L Align, and DM refer to BLEU score, ROUGE score, Multimodal-Language Align, and description mapper, respectively.}
\label{tbl:add}
\vspace{-0.2cm}
\end{table}
\\
\noindent\textbf{Quantitative Evaluation of MLLMs and Prompts.}
We evaluate MLLM-generated SL descriptions.
Details on each MLLM’s characteristics are provided in the supplementary.
First, to assess how effectively SL descriptions represent SL images, we analyze the performance of using only SL descriptions without videos, as shown in Tab.~\ref{tbl:mmlm_type}. 
Consistent with Sec.~\ref{sec:preliminary}, the LLaVA-OneVision 7B model performs best, focusing on SL components. 
We also identify a limitation in using descriptions alone, as indicated by the deficient BLEU-4 score.
Next, we analyze the results obtained using SL descriptions from various MLLMs within the proposed framework.
All models except InternVL2~\cite{chen2024internvlscalingvisionfoundation} outperform the previous gloss-free SOTA, SignLLM~\cite{gong2024llms}, highlighting consistent SLT improvements with SL descriptions.

\begin{table}[h!]
\centering
\scriptsize
\setlength{\tabcolsep}{2pt}
\renewcommand{\arraystretch}{1.1}
    \begin{tabular}{l|cc}
    \hline
    Model & B-4 (w/o visual) & B-4 \\
    \hline
    LLaVA-OneVison 0.5B~\cite{li2024llavaov}                          &  6.53   &  24.96    \\
    LLaVA-NeXT 7B~\cite{li2024llavanext}                              &  5.76   &  25.01      \\
    LLaVA-OneVison 7B~\cite{li2024llavaov}                            &  \textbf{8.99}   &  \textbf{25.73}  \\
    InternVL2 8B~\cite{chen2024internvlscalingvisionfoundation}       &  5.93   &  23.16    \\
    Qwen2VL 8B~\cite{wang2024qwen2vlenhancingvisionlanguagemodels}    &  6.40   &  25.37    \\
    Pixtral 12B~\cite{agrawal2024pixtral12b}                          &  6.35   &  24.58  \\ \hline
    \end{tabular}
    \caption{Performance comparison using various MLLMs. B-4 (w/o visual) indicates the BLEU-n score obtained when translations are performed using only SL descriptions.
    }
\label{tbl:mmlm_type}
\vspace{-.1cm}
\end{table}
We also compare different prompts.
For this, we select prompts (2), (3), and (5) from each group—simple (1, 2), detailed (3, 4), and in-context (5, 6) in Fig.~\ref{fig:prompt}—based on how accurately and thoroughly they describe SL components.
As shown in Tab.~\ref{tbl:prompt eng}, consistent with the qualitative results (Sec.~\ref{sec:preliminary}), including more accurate and detailed SL components improves SLT performance.
\begin{table}[h!]
\centering
\scriptsize
\setlength{\tabcolsep}{2pt}
\renewcommand{\arraystretch}{1.1}
\begin{tabular}{c|cccc}
    \hline
                  Prompts & B-1 & B-2 & B-3 & B-4 \\ 
                  \hline
                (2) &    48.21    &   37.61    &    30.20    &    25.00    \\
                (3) &    \textbf{48.92}   &   \textbf{38.12}   &  \textbf{30.79}   &  \textbf{25.73}  \\
                (5) &    46.68    &    35.75    &   28.17     &    23.10   \\ \hline
\end{tabular}
\caption{Performance comparison based on various prompts.}
\label{tbl:prompt eng}
\vspace{-.1cm}
\end{table}

\noindent\textbf{Description Encoder.}
Accurately embedding SL descriptions without distorting their meaning is crucial.
To identify the most effective embedding features for MMSLT, we compare the translation performance of several sentence-level embedding models: BERT~\cite{devlin2019bertpretrainingdeepbidirectional} and SentenceBERT~\cite{reimers-2019-sentence-bert}, the text encoder mBART~\cite{tang2020multilingualtranslationextensiblemultilingual}, and the LLM Qwen2~\cite{tang2020multilingualtranslationextensiblemultilingual} as description encoders.
Following previous works~\cite{izacard2022unsupervised, wang2024text}, we use token embeddings averaged by mean pooling for mBART and Qwen2.
Contrary to our expectations, Tab.~\ref{tbl:de} demonstrates that using the BERT~\cite{devlin2019bertpretrainingdeepbidirectional} model yields the best performance.
This result can be attributed to BERT's capability to capture detailed context through sentence-level embeddings. In contrast, SentenceBERT, by focusing on the relationships between pair of sentences, may fail to adequately capture the fine-grained context of individual sentences. Additionally, mBART and Qwen2 may lose critical temporal information due to mean pooling. In particular, the large embedding dimension (3,584) in Qwen2 can result in the loss of visual information during the integration process. 
All encoders, except Qwen2, surpass SignLLM~\cite{gong2024llms} in Tab.~\ref{tbl:phoenix} by a large margin, underscoring the superior effectiveness of our model's design.\\
\begin{table}[h!]
\vspace{-.4cm}
\centering
\scriptsize
\renewcommand{\arraystretch}{1.1}
    \begin{tabular}{l|cccc}
    \hline
    Model   & B-1 & B-2 & B-3 & B-4 \\
                  \hline
    BERT~\cite{devlin2019bertpretrainingdeepbidirectional}   &   \textbf{48.92}   &    \textbf{38.12}    &    \textbf{30.79}     &   \textbf{25.73}     \\
    SentenceBERT~\cite{reimers-2019-sentence-bert}   & 48.69    &   37.84     &    30.50    &    25.40    \\
    mBART$^\dagger$~\cite{tang2020multilingualtranslationextensiblemultilingual}  &    48.40    &    37.43    &   29.68    &    24.44    \\
    Qwen2$^\dagger$~\cite{yang2024qwen2technicalreport}   &  45.94    &    34.97    &    27.55    &    22.64    \\ \hline
    \end{tabular}
    \caption{Ablation study on $\psi_{de}$. $\dagger$ uses token embeddings averaged by mean pooling.
}
\label{tbl:de}
\vspace{-0.1cm}
\end{table}

\subsection{Qualitative Results}
Tab.~\ref{tbl:qualitative} presents the translation results for randomly selected SL videos from the PHOENIX14T test set.
We visualize the reference, GFSLT-VLP~\cite{zhou2023gloss} (the only other gloss-free model with publicly available code), and MMSLT.
The results indicate that while GFSLT-VLP correctly translates only a few words or generates entirely different sentences, MMSLT performs more competently, often replacing certain words with others of similar meaning while preserving the overall meaning.
These examples qualitatively show MMSLT's translation effectiveness.

\begin{table}[ht]
\centering
    \tiny
    \setlength{\tabcolsep}{1pt}
    \renewcommand{\arraystretch}{1.1}
    \begin{tabular}{ll}
    \hline
    \textbf{Reference}: & \begin{tabular}[c]{@{}l@{}}am sonntag im norden und in der mitte schauer dabei ist es im norden stürmisch\\ (On Sunday there will be showers in the north and in the centre, although it will be stormy in the north)\end{tabular}\\
    \textbf{GFSLT-VLP}: & \begin{tabular}[c]{@{}l@{}}\textcolor{blue}{am sonntag} \textcolor{red}{in der nordhälfte regenwolken und stürmisch} \textcolor{blue}{im norden ist es} \textcolor{red}{weiter} \textcolor{blue}{stürmisch}\\ (\textcolor{blue}{On Sunday there will be} \textcolor{red}{rain clouds and storms in the northern half, but it will continue to be} \\ \textcolor{blue}{stormy in the north})\end{tabular}\\
    \textbf{MMSLT}:     & \begin{tabular}[c]{@{}l@{}}\textcolor{blue}{am sonntag im norden und} \textcolor{red}{süden} \textcolor{green}{regenschauer} \textcolor{blue}{im norden} \textcolor{red}{mitunter} \textcolor{blue}{stürmisch}\\ (\textcolor{blue}{On Sunday there will be showers in the north} \textcolor{red}{and south,} \textcolor{red}{sometimes} \textcolor{green}{stormy in the north})\end{tabular}\\ \hline
    \textbf{Reference}:  & \begin{tabular}[c]{@{}l@{}}örtlich schauer oder gewitter die heftig sein können\\ (Local showers or thunderstorms which can be heavy)\end{tabular}\\
    \textbf{GFSLT-VLP}: & \begin{tabular}[c]{@{}l@{}}\textcolor{red}{bei schauern und} \textcolor{green}{gewittern} \textcolor{red}{starke böen}\\ (\textcolor{red}{Strong gusts during} \textcolor{green}{showers} \textcolor{red}{and thunderstorms})\end{tabular}\\
    \textbf{MMSLT}:     & \begin{tabular}[c]{@{}l@{}}\textcolor{green}{hier und da gibt es} \textcolor{blue}{schauer oder gewitter die heftig sein können}\\ (\textcolor{green}{Here and there are} \textcolor{blue}{showers or thunderstorms which can be heavy})\end{tabular}\\ \hline
    \textbf{Reference}: &  \begin{tabular}[c]{@{}l@{}}der wind weht meist schwach und kommt aus unterschiedlichen richtungen\\ (The wind is usually weak and comes from different directions)\end{tabular}\\
    \textbf{GFSLT-VLP}: &  \begin{tabular}[c]{@{}l@{}}\textcolor{red}{schwacher bis mäßiger} \textcolor{blue}{wind aus unterschiedlichen richtungen}\\ (\textcolor{red}{Light to moderate} \textcolor{blue}{winds from different directions})\end{tabular}\\
    \textbf{MMSLT}:     &  \begin{tabular}[c]{@{}l@{}}\textcolor{blue}{der wind weht meist schwach aus unterschiedlichen richtungen}\\ (\textcolor{blue}{The wind usually blows weakly from different directions})\end{tabular}\\ \hline
    \textbf{Reference}: &  \begin{tabular}[c]{@{}l@{}}am freitag scheint abseits der nebelgebiete häufig die sonne\\ (On Friday the sun often shines outside of the foggy areas)\end{tabular}\\
    \textbf{GFSLT-VLP}: &  \begin{tabular}[c]{@{}l@{}}\textcolor{blue}{am freitag} \textcolor{red}{sobald} \textcolor{red}{der nebel weg ist sonnenschein}\\ (\textcolor{blue}{On Friday} \textcolor{red}{as soon as the fog is gone there will be sunshine})\end{tabular}\\
    \textbf{MMSLT}:     &  \begin{tabular}[c]{@{}l@{}}\textcolor{blue}{am freitag scheint abseits der nebelfelder häufig die sonne}\\ (\textcolor{blue}{On Friday the sun often shines outside of the foggy areas})\end{tabular}\\ \hline
    \end{tabular}
    \caption{Translation results. 
    Correct answers are marked in blue, semantically similar but rephrased answers are displayed in green, and incorrect answers are indicated in red.
    }
    \label{tbl:qualitative}
    \vspace{-0.1cm}
\end{table}

\section{Conclusion}
\label{sec:conclusion}
In this study, we propose MMSLT, a gloss-free SLT framework that, for the first time, leverages an off-the-shelf MLLM.
We utilize the MLLM to generate SL descriptions through carefully crafted prompts, integrating them with SL images to effectively represent SL. 
We introduce a pre-training module to align the fused modalities with spoken sentences, addressing modality gaps in SLT. 
Additionally, we propose a description mapper to reduce the computational burden of the MLLM during inference by approximating the SL descriptions. 
This research not only lays the foundation for using MLLMs in SLT but also paves the way for future advancements in multimodal learning across various computer vision fields.
\clearpage
\section{Acknowledgement}
This work was supported by the National Research Foundation (NRF) grants funded by Korean Government (MSIT) (No. 2023R1A2C200337911 and No. RS-2023-00220762) and the Institute of Information \& Communications Technology Planning \& Evaluation (IITP) grant funded by the Korea government (MSIT) (No.RS-2025-02303870, Software Technology for Efficient Multimodal Visual Information Processing in High-Speed Spatial Interactions).

{
    \small
    \bibliographystyle{ieeenat_fullname}

}

\end{document}